\documentclass[sigconf]{acmart} 
\AtBeginDocument{%
  }
\usepackage{multirow}
\usepackage{makecell}
\usepackage{xcolor}
\definecolor{mygreen}{RGB}{112,173,71}
\definecolor{myorange}{RGB}{192,97,21}
\setcopyright{acmlicensed}
\copyrightyear{2026}
\acmYear{2026}
\acmConference[CIKM '26]{35th International ACM Conference on Information and Knowledge Management}{November 7-11,
  2026}{Rome, ITALY}

\begin{document}

\title{OmniHandwritingOCR: A Diagnostic Benchmark for Evaluating Multimodal LLMs in Handwritten OCR Scenarios}


\author{Zinuo Guo}
\email{51275901001@stu.ecnu.edu.cn}
\affiliation{%
  \institution{East China Normal University}
  \city{Shanghai}
  \country{China}
}
\author{Min Zhang}
\email{mzhang@cs.ecnu.edu.cn}
\authornote{Corresponding author \& Project leader.}
\affiliation{%
  \institution{East China Normal University}
  \city{Shanghai}
  \country{China}
}
\author{Bo Jiang}
\email{bjiang@deit.ecnu.edu.cn}
\affiliation{%
  \institution{East China Normal University}
  \city{Shanghai}
  \country{China}
}


\begin{abstract}
  Multimodal large language models (MLLMs) are increasingly used as OCR systems in document and knowledge-processing pipelines, but their ability to faithfully read real handwriting remains underexplored. Existing OCR benchmarks focus largely on printed text or clean single-line inputs, leaving limited coverage of realistic handwritten OCR scenarios such as multilingual handwriting, writer errors, and structurally complex mathematical expressions. We introduce \textbf{OmniHandwritingOCR}, a diagnostic benchmark for evaluating MLLMs and OCR systems on handwritten OCR. It covers handwritten text recognition and handwritten mathematical expression recognition across six subtasks and twelve subsets, totaling 77.57K labeled images from public datasets and newly collected student writings. A key component is a difficulty-stratified multi-line formula corpus designed to test robustness under increasing structural complexity. We evaluate thirteen open- and closed-source systems with five complementary metrics under a unified protocol. Results show that current systems remain far from faithful transcription: performance drops sharply on complex multi-line formulas, model rankings vary across language and formula settings, and several generative models hallucinate plausible but visually unsupported corrections. OmniHandwritingOCR provides a challenging testbed for diagnosing language, content, structural, and visual-grounding failure modes of multimodal models in handwritten OCR scenarios. The code is available at \url{https://github.com/ECNU-RAIL/OmniHandwritingOCR-CIKM2026}.
\end{abstract}



\begin{teaserfigure}
  \centering
  \includegraphics[width=0.89\linewidth]{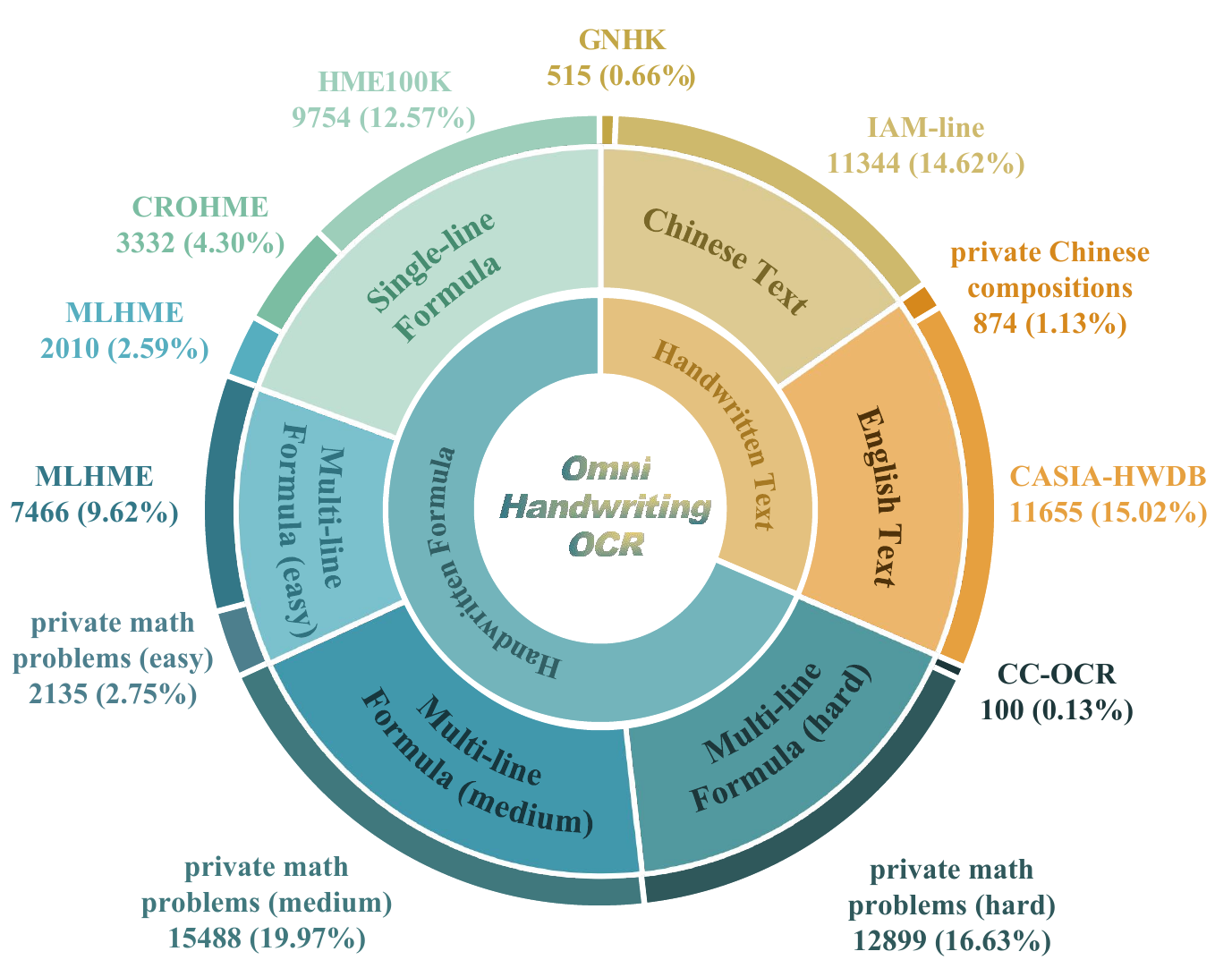}
  \caption{Overview of OmniHandwritingOCR. The benchmark focuses on two core tasks, handwritten text recognition (HTR) and handwritten mathematical expression recognition (HMER), covers six subtasks and twelve subsets with 77.57K images.}
  \label{fig:overview}
\end{teaserfigure}

\maketitle

\section{Introduction}
Optical character recognition (OCR) is a basic component of information acquisition and knowledge processing pipelines. In information and knowledge management settings, OCR outputs often become the input to indexing, search, retrieval-augmented generation, educational analytics, and document-level knowledge extraction systems. Modern systems are no longer limited to specialized OCR engines: multimodal large language models (MLLMs)~\cite{haotian_visual_2023,gpt4v,bai_qwen-vl_2023,deepseekvl} are now used to read documents, extract knowledge from images, and convert visual content into structured text. While these models have achieved strong performance on printed documents, their reliability on handwriting remains unclear. Real handwritten content contains blurred strokes, corrections, irregular layouts, writer-specific styles, and incomplete or erroneous expressions. These properties are exactly where a generative model may stop recognizing visual evidence and start producing plausible but unsupported text.

Handwriting recognition is commonly studied through two related tasks: handwritten text recognition (HTR)~\cite{lstm,graves_offline_2008,kumar2023automated}, which transcribes natural-language handwriting, and handwritten mathematical expression recognition (HMER)~\cite{syntax_1967,online_hmer_2014,zhao2021handwrittenmathematicalexpressionrecognition}, which converts two-dimensional formula structures into markup such as \LaTeX. Existing resources have advanced both directions, but they leave important evaluation gaps for current MLLM/OCR systems. Classical HTR datasets such as IAM~\cite{marti_iam-database_2002} and RIMES~\cite{grosicki_results_2009} are largely monolingual and often cleaner than real educational writing. HMER datasets such as CROHME~\cite{fink_icdar_2023} and MathWriting~\cite{gervais_mathwriting_2025} have driven progress on mathematical transcription, but are dominated by single-line expressions. As a result, existing evaluations do not sufficiently test multilingual handwriting, real student work, fact-preserving transcription, or complex multi-line formula structures.

This limitation is not only a dataset coverage issue; it is an evaluation issue. For LLM-based OCR systems, a useful benchmark should expose where models fail, how performance changes with structural complexity, and when outputs drift from visual evidence into plausible correction. Aggregate OCR scores can hide these failures, especially when a model appears fluent while omitting symbols, repairing student mistakes, or fabricating content that is not present in the image. Such errors directly affect downstream information and knowledge management applications, where faithful transcription is often more important than plausible reconstruction.

To address these needs, we introduce \textbf{OmniHandwritingOCR}, a diagnostic benchmark for evaluating handwritten OCR in MLLM/\-OCR systems. It covers English text, Chinese text, single-line formulas, and multi-line formulas stratified into easy, medium, and hard subsets, totaling 77.57K image-label pairs from public and newly collected sources. Its diagnostic design supports analysis along four axes: language, content type, structural complexity, and visual grounding. This allows the benchmark to report not only which system ranks highest overall, but also where and why a model fails under multilingual, long-context, and structurally complex handwritten inputs.

Our contributions are as follows:
\begin{itemize}
    \item We construct a large-scale handwritten OCR benchmark that integrates HTR and HMER, covering six subtasks and twelve subsets with 77.57K labeled images.
    \item We introduce a difficulty-stratified multi-line formula evaluation setting that targets complex real student work and enables analysis of model degradation as structural complexity increases.
    \item We define a unified evaluation setting with task-specific normalization and tokenization rules, enabling fair comparison across general-purpose MLLMs and specialized OCR models.
    \item We benchmark thirteen open- and closed-source systems and identify diagnostic failure modes, including language-specific performance shifts, sensitivity to formula structure, and hallucinated corrections on ambiguous handwriting.
\end{itemize}

\section{Related Work}
\label{sec:related_work}

We review prior work from the perspective of handwritten recognition and benchmark coverage, focusing on handwritten text recognition (HTR), handwritten mathematical expression recognition (HMER), and the gap between existing resources and realistic handwritten OCR scenarios for current MLLM/OCR systems.

\begin{figure*}[!t]
  \centering
  \includegraphics[width=\textwidth]{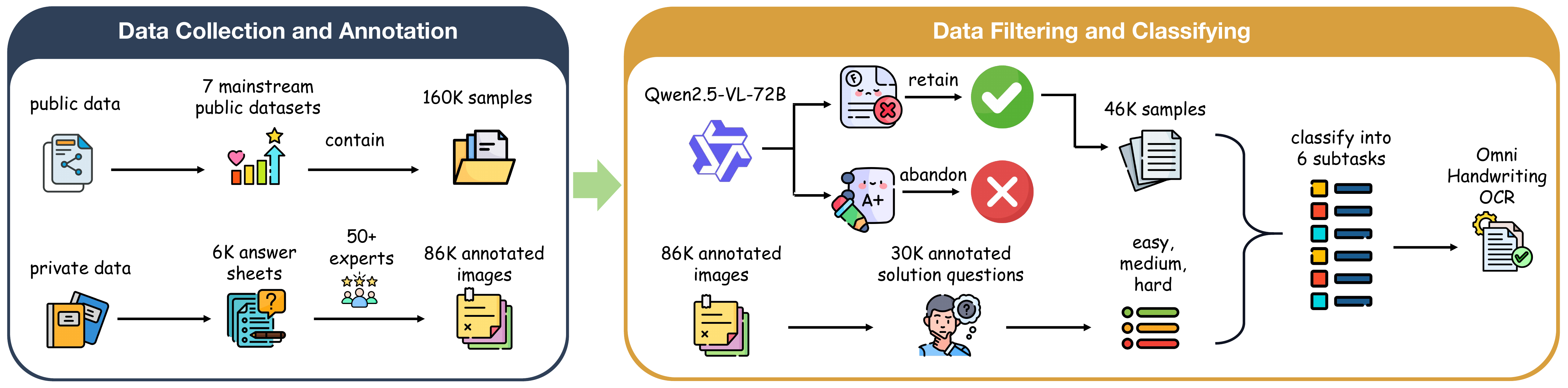}
  \caption{Construction pipeline of OmniHandwritingOCR, including data collection, filtering, difficulty stratification, annotation, and expert verification.}
  \label{fig:pipeline}
\end{figure*}

\subsection{Handwritten Text Recognition}
\label{sec:handwritten_text_recognition}

Handwritten text recognition (HTR) has long been studied as a core document analysis problem. Earlier systems relied on statistical sequence modeling, including HMM-based recognizers~\cite{hmm}, while modern systems commonly use CNN-RNN-CTC architectures~\cite{shi_end--end_2015}, multidimensional recurrent models~\cite{graves_offline_2008}, attention-based recognizers~\cite{bahdanau_neural_2016}, and Transformer-based models such as TrOCR~\cite{li_trocr_2022}. These methods have substantially improved line-level and document-level transcription, and they remain important baselines for understanding how visual encoders and sequence decoders handle handwritten variation. However, real handwritten text still poses challenges beyond isolated character recognition, including inconsistent spacing, writer-specific abbreviations, touching strokes, insertions, deletions, and layout changes across lines.

Progress in HTR has been driven by public benchmarks such as IAM~\cite{marti_iam-database_2002}, RIMES~\cite{grosicki_results_2009}, READ-BAD~\cite{gruning2017readbadnewdatasetevaluation}, GNHK~\cite{GNHK}, CASIA-HWDB~\cite{liu_casia_2011}, and SCUT-HCCDoc~\cite{scut_hccdoc}. These datasets cover important scenarios including English handwriting, French mail, historical documents, in-the-wild handwriting, and Chinese handwritten text. However, they are often organized around specific languages, sources, or document types, and many samples are cleaner than real educational handwriting. They also do not directly evaluate how general-purpose MLLMs behave when handwriting contains writer errors, corrections, irregular layouts, or mixed textual and mathematical content.

Recent OCR and document benchmarks for MLLMs, such as OmniDocBench~\cite{ouyang_omnidocbench_2025}, OCRBenchv2~\cite{fu_ocrbench_2025}, and CC-OCR~\cite{yang_cc-ocr_2024}, broaden evaluation to layout-rich, multilingual, and document-level tasks. Nevertheless, handwriting remains only one component within these broader resources. OmniHandwritingOCR follows the HTR tradition but extends it toward MLLM-era evaluation by combining English and Chinese handwriting with fact-based annotations and unified scoring, enabling direct diagnosis of faithful transcription under realistic handwritten conditions.

\subsection{Handwritten Mathematical Expression Recognition}
\label{sec:handwritten_math_recognition}

Handwritten mathematical expression recognition (HMER) is more structurally demanding than ordinary text recognition because a system must identify symbols and recover two-dimensional relations. Classical and neural approaches have explored grammar-based parsing~\cite{syntax_1967}, online trajectory modeling~\cite{online_hmer_2014}, sequence decoders~\cite{zhang_gru-based_2017}, tree-structured decoders~\cite{zhang_tree-structured_2020}, graph modeling~\cite{gnn}, and structure-aware Transformers such as TAMER~\cite{zhu_tamer_2024}. These methods show that formula recognition requires explicit or implicit modeling of spatial relations, not only character-level transcription. In this setting, small recognition errors such as missing braces, misplaced superscripts, or incorrect fraction scopes can substantially change the recovered expression, making structural fidelity as important as symbol accuracy.

Benchmarks such as CROHME~\cite{fink_icdar_2023}, HME100K~\cite{yuan_syntax-aware_2022}, MathWriting~\cite{gervais_mathwriting_2025}, UniMER~\cite{unimernet}, and MLHME~\cite{MLHME} are central to HMER research. They provide standardized evaluation data and have enabled progress on symbol recognition, \LaTeX{} sequence generation, and expression-level parsing. However, much of the existing evaluation is still dominated by single-line expressions or relatively controlled formula layouts. Even when multi-line formulas are included, they usually do not fully capture the long, noisy, and correction-heavy derivations found in real student answer sheets.

This gap is important for current MLLM/OCR systems. A model may recognize isolated symbols or short formulas while failing to preserve alignment, multi-step dependencies, cases, fractions, superscripts, or crossed-out content in longer handwritten derivations. OmniHandwritingOCR therefore integrates HTR and HMER in a single benchmark and places special emphasis on difficulty-stratified multi-line formulas. By combining public formula datasets with newly collected student answer sheets, it evaluates not only formula transcription accuracy, but also robustness to the structural complexity that appears in realistic handwritten mathematical work.

Compared with prior HTR or HMER resources, our focus is not to replace task-specific benchmarks, but to provide a unified stress test for systems that are increasingly deployed as general handwritten OCR engines. This is particularly relevant for MLLMs, whose outputs may be influenced by language priors, mathematical priors, and instruction-following behavior. Evaluating text and formula handwriting under the same protocol makes these cross-task effects visible and helps distinguish genuine visual recognition from plausible reconstruction.

\begin{figure*}[htbp]
  \centering
  \includegraphics[width=\textwidth]{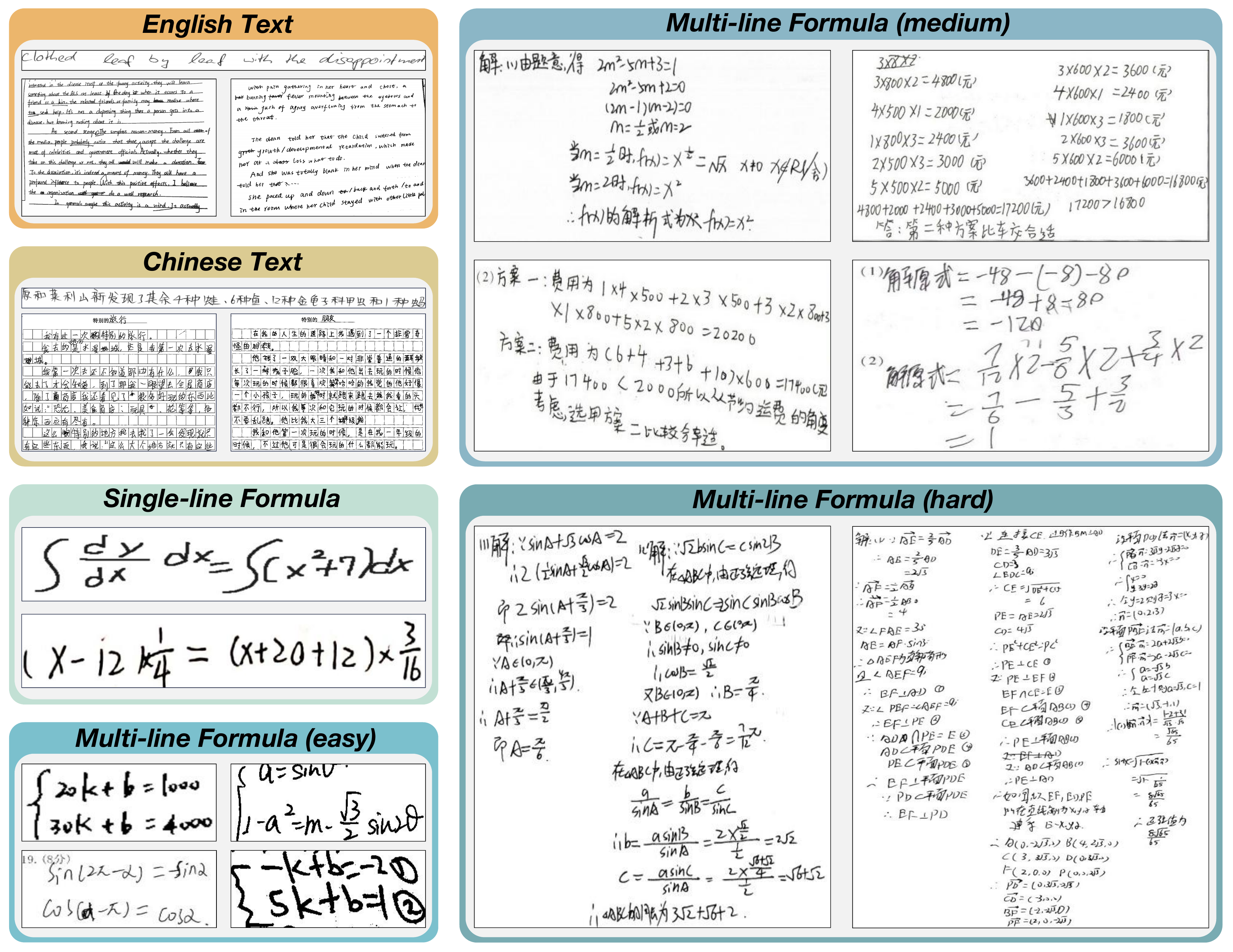}
  \caption{Representative examples from each subtask, showing the diversity of handwriting styles and the progressive difficulty of formula structures.}
  \label{fig:cases}
\end{figure*}

\begin{figure}[htbp]
  \centering
  \includegraphics[width=\linewidth]{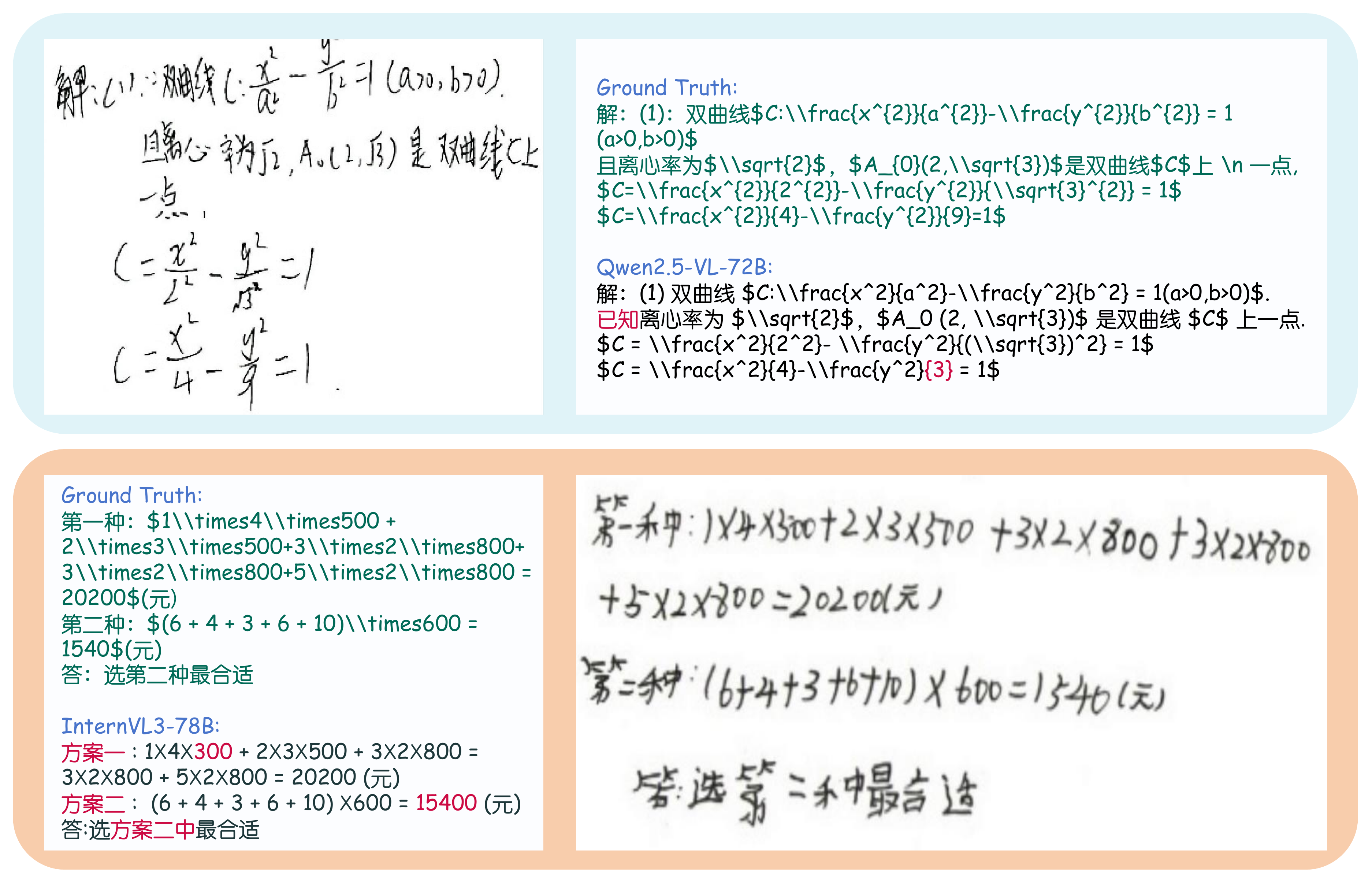}
  \caption{Examples of fact-based annotations. Ground-truth labels preserve visible writer mistakes, while Qwen2.5-VL-72B and InternVL3-78B incorrectly produce corrected but visually unsupported transcriptions.}
  \label{fig:label}
\end{figure}

\begin{table}
    \centering
    \small
    \caption{Detailed composition of the dataset.}
    \label{tab:dataset_composition}
    \begin{tabular}{lccc}
    \toprule
    \textbf{Subtask} & \textbf{Source} & \textbf{Samples} & \textbf{Total} \\
    \midrule
    \multirow{2}{*}{English Text} & GNHK & 515 & \multirow{2}{*}{11,859} \\
                             & IAM-line & 11,344 & \\
    \midrule
    \multirow{2}{*}{Chinese Text} & private & 874 & \multirow{2}{*}{12,529} \\
                             & CASIA-HWDB  & 11,655 & \\
    \midrule
    \multirow{3}{*}{\makecell[l]{Single-line\\Formula}} & CROHME & 3,332 & \multirow{3}{*}{15,096} \\
                        & HME100K & 9,754 & \\
                        & MLHME & 2,010 & \\
    \midrule
    \multirow{2}{*}{\makecell[l]{Multi-line\\Formula (easy)}} & MLHME & 7,466 & \multirow{2}{*}{9,601} \\
                             & private & 2,135 & \\
    \midrule
    \makecell[l]{Multi-line\\Formula (medium)} & private & 15,488 & 15,488 \\
    \midrule
    \multirow{2}{*}{\makecell[l]{Multi-line\\Formula (hard)}} & CC-OCR & 100 & \multirow{2}{*}{12,999} \\
                             & private & 12,899 & \\
    \midrule
    \textbf{Total} & & & \textbf{77,572} \\
    \bottomrule
    \end{tabular}
\end{table}

\section{OmniHandwritingOCR}
\label{sec:benchmark}

To evaluate handwriting recognition capabilities beyond aggregate OCR scores, we design OmniHandwritingOCR around three requirements: broad task coverage, difficulty-aware diagnosis, and standardized scoring. The benchmark contains more than 77K samples, spans English text, Chinese text, and mathematical formulas, and systematically stratifies multi-line formula content by difficulty. This dual focus on breadth and depth enables fine-grained analysis of model generalization across languages, layouts, and structural complexity.

\subsection{Design Principles}
OmniHandwritingOCR is designed to support diagnostic evaluation rather than only leaderboard ranking. The first principle is \textit{task diversity}. Real handwritten OCR applications rarely involve a single clean text line. They may contain English notes, Chinese compositions, mathematical expressions, long derivations, corrections, and mixed natural-language/formula content. We therefore combine HTR and HMER in one benchmark so that models can be compared under a shared protocol while still exposing task-specific strengths and weaknesses.

The second principle is \textit{difficulty-aware structure}. For formula recognition, the central challenge is not only identifying symbols, but also preserving two-dimensional relations such as fractions, superscripts, cases, and aligned multi-step derivations. A benchmark dominated by single-line expressions can overestimate model readiness for real educational or scientific handwriting. OmniHandwritingOCR therefore separates single-line formulas from multi-line formulas and further stratifies multi-line samples by difficulty, enabling controlled analysis of how models degrade as visual and symbolic structure becomes more complex.

The third principle is \textit{faithful transcription}. In many knowledge-processing pipelines, OCR output is treated as evidence for indexing, retrieval, or downstream reasoning. A model that corrects a student's mistake or fills in a missing derivation may produce a plausible answer, but it is no longer faithfully transcribing the image. Our fact-based annotation policy and unified scoring protocol are designed to penalize such visually unsupported corrections. This makes the benchmark especially suited to evaluating generative OCR systems, where hallucinated correction can be hidden by fluent output.

\subsection{Benchmark Construction}
In this section, we provide a detailed overview of our construction pipeline for OmniHandwritingOCR, as shown in Figure \ref{fig:pipeline}. The pipeline includes dataset collection to ensure breadth and novelty, data filtering and difficulty stratification to ensure evaluative rigor, and finally, expert annotation to ensure ground-truth accuracy.

\subsubsection{Dataset Collection}
Our data collection strategy employs a two-pronged approach. First, we aggregated key public datasets to establish a broad foundation. We then supplemented this with a substantial, newly collected private dataset to target specific and challenging domains.

\textbf{Public Data Aggregation.} To establish a comprehensive foundation, we gathered samples from influential public datasets, including IAM-line \cite{marti_iam-database_2002} and GNHK \cite{GNHK} for English text, CASIA-HWDB \cite{liu_casia_2011} for Chinese text, and CROHME \cite{fink_icdar_2023}, HME100K \cite{yuan_syntax-aware_2022}, MLHME38K \cite{MLHME}, and CC-OCR \cite{yang_cc-ocr_2024} for mathematical formulas. This initial step provided a diverse pool of approximately 160K samples, covering a wide range of established tasks.

\textbf{Private Data Collection.} Our private data is collected specifically to address the most critical gaps in existing resources. We sourced it from two key areas: (1) 874 authentic student Chinese language compositions, which fill a crucial void by providing complex, real-world multi-line samples, and (2) our primary contribution: a first-of-its-kind dataset derived from over 6,000 authentic student mathematics answer sheets. From these, we segmented and annotated 86K images. This unique corpus captures the natural, often chaotic, and structurally complex nature of handwritten problem-solving, a scenario largely absent from existing datasets that are dominated by clean, single-line expressions.

\subsubsection{Data Filtering and Difficulty Stratification}
A primary goal was to create a benchmark that actively challenges state-of-the-art models, rather than just measuring performance on solved problems. To this end, we implemented a rigorous, two-part filtering and stratification process.

First, for the public data, we performed a challenge-oriented filtering to increase difficulty. We leveraged a powerful vision-language model, Qwen2.5-VL-72B, to recognize the initial 160K samples. By calculating the normalized edit distance (NED) between the model's prediction and the ground truth, we were able to systematically identify and discard samples that the model found easy. This deliberate selection of poorly performing samples allowed us to distill a final set of 46K high-quality, challenging samples that push the boundaries of current model capabilities.

Second, for the private math data, we first focused the dataset by filtering the 86K segmented images to retain only the core solution sections, resulting in 30K high-quality samples of multi-line and difficult formulas. Furthermore, to enable detailed model evaluation, we designed a quantitative difficulty scoring system. This system scores each sample with three factors: the length of the ground truth (weight 0.2, representing information volume), the number of complex \LaTeX\ commands (weight 0.3, representing structural complexity), and the recognition NED from GPT-4o (weight 0.5, representing perceptual difficulty for a strong vision-language model). We use this score only to stratify evaluation subsets, not as a training signal or as an additional metric. Because one component is model-informed, we validate the resulting split with model-agnostic structural statistics below. Based on this composite score, we classify the samples into easy, medium, and hard categories, forming the hierarchical structure that supports the diagnostic experiments in Section~\ref{sec:experiments}.

To reduce the risk that the difficulty split is merely tied to one model's behavior, we also inspect model-agnostic label statistics. As shown in Table~\ref{tab:difficulty_validation}, the three private multi-line formula subsets exhibit monotonic increases in ground-truth length, \LaTeX\ command density, and number of non-empty lines. This validates that the split reflects observable structural complexity in addition to model recognition difficulty.

\begin{table}[tbp]
    \centering
    \small
    \caption{Model-agnostic statistics of private multi-line formula subsets. Len. denotes average ground-truth characters, Cmd. denotes average \LaTeX\ commands, and Lines denotes average non-empty text lines.}
    \vspace{-2mm}
    \label{tab:difficulty_validation}
    \begin{tabular}{lccc}
    \toprule
    \textbf{Subset} & \textbf{Len.} & \textbf{Cmd.} & \textbf{Lines} \\
    \midrule
    ml-easy & 156.9 & 4.1 & 3.7 \\
    ml-medium & 213.0 & 7.6 & 7.1 \\
    ml-hard & 488.8 & 25.7 & 13.0 \\
    \bottomrule
    \end{tabular}
\end{table}

\subsubsection{Expert Annotation and Verification}
Precise and high-quality annotation is critical for a fair and research-worthy benchmark. Our annotation strategy was tailored to the data's origin to ensure consistency and accuracy. 
For all public data, we retained their original, widely-used annotations to maintain consistency for comparison with existing research and also honor the significant community effort invested in creating these established ground truths.

The annotation of our private data, however, demanded a substantial effort. We assembled a dedicated team of over 50 domain experts for an intensive two-month project. The final benchmark contains 31,396 private samples, including 874 Chinese composition samples and 30,522 private mathematical-expression samples. The private mathematical-expression data were derived from over 6,000 authentic student answer sheets, and each original answer sheet underwent two rounds of human checking before its segmented samples were finalized. In the first round, a human expert verified and corrected the model-generated pre-annotations. The human-refined results then served as higher-quality baselines for a second round of expert review and refinement. Ambiguous cases were resolved by additional inspection rather than majority voting, because many errors involve fine visual details such as superscripts, omitted strokes, or crossed-out content. This iterative process was designed to systematically minimize errors and ensure the final ground truth achieved the highest possible level of accuracy.

A cornerstone of this process was our fact-based annotation principle: maintaining strict fidelity to the writer's original handwriting. We intentionally retained common writer errors, such as misspellings or miscopied numbers, rather than correcting them. Furthermore, any content crossed out by the writer was omitted entirely. This philosophy is crucial because it forces models to learn robust visual pattern recognition instead of relying on contextual guessing or hallucinating corrections, as shown in Figure \ref{fig:label}. By creating a ground truth that reflects what is visually present, our benchmark provides a more accurate tool for pinpointing specific model weaknesses and driving future improvements in genuine recognition capabilities.

\textbf{Privacy and ethical handling.} The private portion consists of educational handwriting samples. Before benchmark construction, personally identifying metadata was removed, and images were segmented to retain only task-relevant handwritten content. The released benchmark uses anonymized identifiers and excludes information that is not needed for recognition evaluation.

\begin{table*}[htbp]
    \centering
    \caption{Comparison of different handwriting-related benchmarks, where HTR denotes Handwritten Text Recognition, HMER denotes Handwritten Mathematical Expression Recognition, and MLF denotes Handwritten Multi-Line Formula.}
    \vspace{-3mm}
    \label{tab:handwriting_benchmarks_compare}
    \begin{tabular}{l c c c c c} 
    \toprule
    \multirow{2}{*}{\textbf{Benchmark}} & \multirow{2}{*}{\textbf{Type}} & \multirow{2}{*}{\textbf{Tasks}} & \multirow{2}{*}{\textbf{Structure}} & \multicolumn{2}{c}{\textbf{Num}} \\
    \cmidrule(lr){5-6}
      & & & & \textbf{All} & \textbf{MLF} \\
    \midrule
    RIMES-2011-line~\cite{grosicki_results_2009} & HTR & 1 & single-line only  & 12,104 & - \\
    READ-2016~\cite{readbad} & HTR & 1 & multi-line only  &  30,000 & - \\
    SCUT-HCCDoc~\cite{scut_hccdoc} & HTR & 5 & single-line \& multi-line  & 12,253 & - \\
    CROHME2023~\cite{fink_icdar_2023} & HMER & 1 & single-line only & 13,279 & 0 \\
    MLHME-38K~\cite{MLHME} & HMER & 2 & single-line \& multi-line & 38,000 & 9,931 \\
    Mathwriting~\cite{gervais_mathwriting_2025} & HMER & 1 & single-line only & 649,000 & 0 \\
    UniMER~\cite{unimernet} & HMER & 4 & single-line \& multi-line & 1,085,548 & 0 \\
    \textbf{Ours} & HTR+HMER & 6 & single-line \& multi-line & 77,572 & 38,088 \\
    \bottomrule
    \end{tabular}
\end{table*}

\subsection{Dataset Statistics}
OmniHandwritingOCR is a large-scale collection totaling 77,572 image-label pairs across 6 different subtasks. The detailed composition, as shown in Table \ref{tab:dataset_composition}, is balanced to cover a wide range of handwritten recognition tasks, from simple text lines to complex, multi-line mathematical expressions.

To elaborate, the English text (en-text) combines 515 images from GNHK and 11,344 from IAM-line. The Chinese text (zh-text) includes 11,655 images from CASIA-HWDB, significantly supplemented by our 874 privately collected multi-line Chinese compositions. A core contribution of our work lies in the mathematical formula portion. The benchmark includes 15,096 single-line (sl) formulas sourced from CROHME (3,332), HME100K (9,754), and MLHME (2,010). More critically, it features our large, unique corpus of multi-line (ml) formulas, which are stratified by our difficulty scoring system into 9,601 ml-easy, 15,488 ml-medium, and 12,999 ml-hard samples.

As shown in Table \ref{tab:handwriting_benchmarks_compare}, OmniHandwritingOCR demonstrates clear advantages in both task comprehensiveness and structural complexity. It uniquely combines handwritten text with handwritten mathematical expressions, allowing it to more closely simulate real-world scenarios involving mixed content. Furthermore, our benchmark contains a substantial number of handwritten multi-line formula samples, 38,088 in total, which is significantly more than the other benchmarks compared. This provides a more challenging platform for evaluating a model's ability to process complex spatial layouts.

\begin{table*}[tbp]
    \centering
    \caption{Experiments (\%) on the entire OmniHandwritingOCR. The highest is \textbf{bold}, and the second highest is \underline{underlined}.}
    \vspace{-3mm}
    \label{tab:performance_on_all}
    \resizebox{0.97\textwidth}{!}{
        \begin{tabular}{lcccc|ccc}
            \toprule
            \textbf{Methods} & \textbf{BLEU-4 \textcolor{mygreen}{$\uparrow$}} & \textbf{F1 Score \textcolor{mygreen}{$\uparrow$}} & \textbf{1-NED \textcolor{mygreen}{$\uparrow$}} & \textbf{Overall \textcolor{mygreen}{$\uparrow$} }& \textbf{CER \textcolor{red}{$\downarrow$}} & \textbf{WER \textcolor{red}{$\downarrow$}} & \textbf{Overall \textcolor{red}{$\downarrow$}} \\
            \midrule 
            \multicolumn{8}{c}{\textcolor{myorange}{\textit{Multimodal large language models (MLLMs) }}} \\
            GPT-4o & 39.53 & 67.66 & 59.32 & 55.50 & 48.53 & 50.03 & 49.28 \\
            Qwen2.5-VL-72B & 56.40 & \underline{79.98} & 72.91 & 69.76 & 38.88 & 36.91 & 37.90  \\
            InternVL3-78B & 53.76 & 78.21 & 72.16 & 68.04 & \underline{32.19} & \underline{34.72} & \underline{33.46} \\
            Gemma-3-27B-it & 31.84 & 60.08 & 50.97 & 47.63 & 65.82 & 67.83 & 66.83 \\
            DeepSeek-VL2-27B & 39.57 & 63.06 & 54.90 & 52.51 & 76.91 & 63.49 & 70.20 \\
            Kimi-VL-A3B-Instruct & \underline{58.75} & 78.90 & \underline{73.58} & \underline{70.41} & 38.20 & 45.28 & 41.74 \\
            Qwen3-VL-8B & \textbf{59.24} & \textbf{81.60} & \textbf{75.63} & \textbf{72.16} &  \textbf{30.94} & \textbf{32.94} & \textbf{31.94} \\
            \midrule
            \multicolumn{8}{c}{\textcolor{myorange}{\textit{Optical character recognition (OCR) models}}} \\
            MonkeyOCR-pro-1.2B & 49.37 & 70.60 & \underline{66.52} & \underline{62.16} & \underline{51.54} & \underline{54.11} & \underline{52.83} \\
            MinerU2.5-1.2B & 8.48 & 25.12 & 21.33 & 18.31 & 125.59 & 124.65 & 125.12 \\
            Nanonets-OCR2-3B & \textbf{53.84} & \textbf{74.13} & \textbf{68.42} & \textbf{65.46} & \textbf{42.27} & \textbf{42.77} & \textbf{42.52} \\
            GOT-OCR2 & 29.03 & 54.19 & 48.32 & 43.85 & 84.19 & 77.65 & 80.92 \\
            PaddleOCR-VL & \underline{50.21} & 69.61 & 62.73 & 60.85  & 71.14 & 72.38 & 71.76 \\
            DeepSeek-OCR & 47.51 & \underline{71.96} & 64.27 & 61.25 & 73.84 & 69.94 & 71.89 \\
            \bottomrule
        \end{tabular}}
\end{table*}

\begin{table*}[tbp]
    \centering
    \caption{Experiments (\%) on the English (en-text) and Chinese text (zh-text). The highest is \textbf{bold}, and the second highest is \underline{underlined}.}
    \vspace{-3mm}
    \label{tab:performance_on_en_zh}
    \resizebox{0.99\textwidth}{!}{
        \begin{tabular}{lcccccc |cccc}
            \toprule
            \multirow{2}{*}{\textbf{Methods}} & \multicolumn{2}{c}{\textbf{BLEU-4 \textcolor{mygreen}{$\uparrow$}}} & \multicolumn{2}{c}{\textbf{F1 Score \textcolor{mygreen}{$\uparrow$}}} & \multicolumn{2}{c}{\textbf{1-NED \textcolor{mygreen}{$\uparrow$}}} & \multicolumn{2}{c}{\textbf{CER \textcolor{red}{$\downarrow$}}} & \multicolumn{2}{c}{\textbf{WER \textcolor{red}{$\downarrow$}}} \\
            \cmidrule{2-11}
            & en-text & zh-text & en-text & zh-text & en-text & zh-text & en-text & zh-text & en-text & zh-text \\
            \midrule 
            \multicolumn{11}{c}{\textcolor{myorange}{\textit{Multimodal large language models (MLLMs) }}} \\
            GPT-4o & 54.03 & 33.51 & 77.44 & 58.48 & 81.29 & 60.66 & 20.05 & 43.90 & 31.93 & 51.84 \\
            Qwen2.5-VL-72B & \textbf{58.57} & \underline{71.51} & \textbf{81.84} & \underline{88.39} & 84.63 & \underline{90.71} & \textbf{16.01} & \textbf{10.42} & \textbf{28.21} & \textbf{15.16} \\
            InternVL3-78B & 51.26 & 56.59 & 76.98 & 77.42 & 81.14 & 79.91 & 20.36 & 22.53 & 34.69 & 29.08 \\
            Gemma-3-27B-it & 40.07 & 13.93 & 69.36 & 33.06 & 74.86 & 29.28 & 26.12 & 95.18 & 42.70 & 101.30 \\
            DeepSeek-VL2-27B & 47.83 & 22.99 & 72.22 & 43.72 & 76.72 & 40.71 & 27.89 & 133.85 & 42.74 & 89.61 \\
            Kimi-VL-A3B-Instruct & 43.65 & 57.53 & 70.77 & 76.73 & 70.77 & 76.15 & 37.96 & 40.34 & 63.22 & 60.03 \\
            Qwen3-VL-8B & \underline{58.37} & \textbf{72.24} & \underline{81.78} & \textbf{88.72} & \textbf{84.76} & \textbf{90.85} & \underline{16.44} & \underline{11.89} & \underline{28.43} & \underline{17.94} \\
            \midrule
            \multicolumn{11}{c}{\textcolor{myorange}{\textit{Optical character recognition (OCR) models}}} \\
            MonkeyOCR-pro-1.2B & \underline{53.46} & 68.09 & \underline{77.82} & 81.83 & \underline{82.41} & 79.66 & \underline{23.64} & 78.67 & \underline{41.78} & 69.15 \\
            MinerU2.5-1.2B & 11.62 & 14.46 & 25.13 & 26.65 & 32.23 & 25.87 & 93.43 & 150.71 & 134.46 & 112.94 \\
            Nanonets-OCR2-3B & \textbf{59.76} & \underline{78.17} & \textbf{81.45} & \underline{89.60} & \textbf{84.96} & \underline{88.25} & \textbf{19.41} & \textbf{22.73} & \textbf{36.79} & \textbf{23.62} \\
            GOT-OCR2 & 21.05 & 24.43 & 47.47 & 48.05 & 59.88 & 50.45 & 43.17 & 62.40 & 70.26 & 69.07 \\
            PaddleOCR-VL & 46.65 & \textbf{81.39} & 68.98 & \textbf{91.65} & 73.66 & \textbf{89.05} & 47.48 & \underline{32.5} & 72.05 & \underline{39.64} \\
            DeepSeek-OCR & 45.75 & 66.77 & 70.14 & 79.48 & 77.33 & 76.95 & 39.83 & 76.06 & 64.09 & 59.14 \\
            \bottomrule
        \end{tabular}}
\end{table*}

\begin{table*}[tbp]
    \centering
    \caption{Experiments (\%) on single-line (sl) and easy multi-line (ml-easy). The highest is \textbf{bold}, and the second highest is \underline{underlined}.}
    \vspace{-3mm}
    \label{tab:performance_on_sl_ml_easy}
    \resizebox{0.99\textwidth}{!}{
        \begin{tabular}{lcccccccccc}
            \toprule
            \multirow{2}{*}{\textbf{Methods}} & \multicolumn{2}{c}{\textbf{BLEU-4 \textcolor{mygreen}{$\uparrow$}}} & \multicolumn{2}{c}{\textbf{F1 Score \textcolor{mygreen}{$\uparrow$}}} & \multicolumn{2}{c}{\textbf{1-NED \textcolor{mygreen}{$\uparrow$}}} & \multicolumn{2}{c}{\textbf{CER \textcolor{red}{$\downarrow$}}} & \multicolumn{2}{c}{\textbf{WER \textcolor{red}{$\downarrow$}}} \\
            \cmidrule{2-11}
            & sl & ml-easy & sl & ml-easy & sl & ml-easy & sl & ml-easy & sl & ml-easy \\
            \midrule 
            \multicolumn{11}{c}{\textcolor{myorange}{\textit{Multimodal large language models (MLLMs) }}} \\
            GPT-4o & 24.51 & 61.08 & 62.26 & 81.68 & 51.98 & 71.17 & 79.24 & 29.37 & 74.50 & 26.35 \\
            Qwen2.5-VL-72B & 40.09 & 64.24 & 72.98 & 82.84 & 63.13 & 71.73 & 67.07 & 35.53 & 57.53 & 30.13 \\
            InternVL3-78B & 56.67 & 67.00 & 84.08 & 85.13 & 80.70 & 75.81 & 27.04 & 26.04 & 30.34 & \underline{23.49} \\
            Gemma-3-27B-it & 19.36 & 52.03 & 53.74 & 75.96 & 45.83 & 64.16 & 104.31 & 40.00 & 108.95 & 35.03 \\
            DeepSeek-VL2-27B & \underline{67.64} & 42.96 & \underline{87.80} & 66.20 & \underline{86.06} & 49.72 & \underline{18.50} & 78.31 & \underline{19.74} & 61.00 \\
            Kimi-VL-A3B-Instruct & \textbf{78.40} & \textbf{84.98} & \textbf{93.33} & \textbf{91.99} & \textbf{93.60} & \textbf{88.82} & \textbf{10.91} & \textbf{21.93} & \textbf{14.20} & \textbf{21.07} \\
            Qwen3-VL-8B & 52.77 & \underline{70.63} & 79.88 & \underline{86.92} & 74.20 & \underline{79.57} & 40.27 & \underline{24.95} & 38.74 & 23.53 \\
            \midrule
            \multicolumn{11}{c}{\textcolor{myorange}{\textit{Optical character recognition (OCR) models}}} \\
            MonkeyOCR-pro-1.2B & \textbf{69.09} & 56.58 & \textbf{88.90} & 71.18 & \textbf{87.56} & \textbf{74.66} & \textbf{16.06} & \textbf{30.39} & \textbf{16.94} & 45.17 \\
            MinerU2.5-1.2B & 5.65 & 2.63 & 24.62 & 16.23 & 24.42 & 11.63 & 132.67 & 123.10 & 107.29 & 108.87 \\
            Nanonets-OCR2-3B & 48.26 & \underline{57.56} & 73.73 & \underline{74.49} & 68.80 & 64.80 & \underline{47.08} & 42.12 & \underline{48.64} & \underline{39.13} \\
            GOT-OCR2 & \underline{56.25} & 28.51 & \underline{78.51} & 56.46 & \underline{74.40} & 39.95 & 57.25 & 83.02 & 52.41 & 70.45 \\
            PaddleOCR-VL & 38.75 & 49.39 & 53.46 & 65.77 & 50.00 & 55.88 & 116.77 & 74.30 & 108.57 & 65.87 \\
            DeepSeek-OCR & 36.97 & \textbf{58.88} & 66.01 & \textbf{79.65} & 55.54 & \underline{68.59} & 154.81 & \underline{38.13} & 118.08 & \textbf{35.50} \\
            \bottomrule
        \end{tabular}}
\end{table*}

\begin{table*}[tbp]
    \centering
    \caption{Experiments (\%) on the medium multi-line formula (ml-medium) and hard multi-line formula (ml-hard). 
    % The highest is \textbf{bold}, and the second highest is \underline{underlined}.
    }
    \vspace{-3mm}
    \label{tab:performance_on_ml_medium_hard}
    \resizebox{0.99\textwidth}{!}{
        \begin{tabular}{lcccccccccc}
            \toprule
            \multirow{2}{*}{\textbf{Methods}} & \multicolumn{2}{c}{\textbf{BLEU-4 \textcolor{mygreen}{$\uparrow$}}} & \multicolumn{2}{c}{\textbf{F1 Score \textcolor{mygreen}{$\uparrow$}}} & \multicolumn{2}{c}{\textbf{1-NED \textcolor{mygreen}{$\uparrow$}}} & \multicolumn{2}{c}{\textbf{CER \textcolor{red}{$\downarrow$}}} & \multicolumn{2}{c}{\textbf{WER \textcolor{red}{$\downarrow$}}} \\
            \cmidrule{2-11}
            & ml-medium & ml-hard & ml-medium & ml-hard & ml-medium & ml-hard & ml-medium & ml-hard & ml-medium & ml-hard \\
            \midrule 
            \multicolumn{11}{c}{\textcolor{myorange}{\textit{Multimodal large language models (MLLMs) }}} \\
            GPT-4o & 46.84 & 17.18 & 73.43 & 52.68 & 61.20 & 29.60 & 43.05 & 75.58 & 43.06 & 72.52 \\
            Qwen2.5-VL-72B & \underline{59.98} & \textbf{43.98} & \underline{80.42} & \textbf{73.39} & \underline{71.47} & \textbf{55.81} & 49.32 & \textbf{54.91} & 42.00 & \textbf{48.42} \\
            InternVL3-78B & 57.53 & 33.52 & 79.35 & 66.32 & 69.81 & 45.59 & \underline{36.39} & 60.80 & \underline{34.83} & 55.89 \\
            Gemma-3-27B-it & 43.61 & 22.03 & 70.32 & 58.05 & 57.32 & 34.38 & 56.11 & 73.21 & 49.95 & 69.05 \\
            DeepSeek-VL2-27B & 37.95 & 18.07 & 62.45 & 45.99 & 48.25 & 27.94 & 109.04 & 93.87 & 81.91 & 85.92 \\
            Kimi-VL-A3B-Instruct & 56.29 & 31.63 & 78.14 & 62.46 & 68.67 & 43.48 & 48.86 & 69.20 & 46.70 & 66.47 \\
            Qwen3-VL-8B & \textbf{61.65} & \underline{39.77} & \textbf{81.82} & \underline{70.45} & \textbf{73.83} & \underline{50.54} & \textbf{34.98} & \underline{57.08} & \textbf{34.45} & \underline{54.52} \\
            \midrule
            \multicolumn{11}{c}{\textcolor{myorange}{\textit{Optical character recognition (OCR) models}}} \\
            MonkeyOCR-pro-1.2B & 26.78 & 22.33 & 51.89 & 53.84 & 40.59 & 33.23 & 83.59 & 76.91 & 77.78 & 73.81 \\
            MinerU2.5-1.2B & 9.98 & 6.53 & 29.28 & 28.83 & 17.82 & 16.23 & 122.57 & 131.06 & 163.02 & 120.28 \\
            Nanonets-OCR2-3B & 48.53 & \underline{30.88} & \underline{72.68} & \textbf{62.80} & \underline{62.04} & 41.67 & \underline{54.71} & \textbf{67.51} & \textbf{46.44} & \textbf{62.44} \\
            GOT-OCR2 & 29.15 & 14.90 & 55.13 & 39.53 & 41.17 & 24.26 & 145.46 & 113.82 & 110.58 & 103.14 \\
            PaddleOCR-VL & \underline{49.68} & \textbf{35.01} & 69.71 & 62.51 & 58.22 & \underline{41.78} & 94.16 & 83.51 & 72.58 & 76.52 \\
            DeepSeek-OCR & \textbf{52.94} & 30.21 & \textbf{75.91} & \underline{61.58} & \textbf{64.96} & \textbf{42.24} & \textbf{53.49} & \underline{70.71} & \underline{48.91} & \underline{65.91} \\
            \bottomrule
        \end{tabular}}
        \vspace{-2mm}
\end{table*}

\section{Experiments}
\label{sec:experiments}

In this section, we evaluate state-of-the-art methods on OmniHandwritingOCR under a unified offline protocol. The goal is not only to rank models, but also to diagnose how their behavior changes across language, handwriting type, and formula complexity.

\subsection{Evaluation Protocol}

\textbf{Baselines.}
We evaluate 13 methods in total, including 7 general-purpose MLLMs and 6 specialized OCR models. To reflect current practice, the benchmark includes both closed-source systems and newly released open-source models. The MLLMs are GPT-4o~\cite{OpenAI2024gpt4osystemcard}, Qwen2.5-VL-72B~\cite{bai2025qwen25vltechnicalreport}, InternVL3-78B~\cite{zhu_internvl3_2025}, Gemma-3-27B-it~\cite{team_gemma_2025}, DeepSeek-VL2-27B~\cite{wu_deepseek-vl2_2024}, Kimi-VL-A3B-Instruct~\cite{team_kimi-vl_2025}, and Qwen3-VL-8B~\cite{Qwen3VL}. The OCR models are MonkeyOCR-pro-1.2B~\cite{li_monkeyocr_2025}, MinerU2.5-1.2B~\cite{wang_mineru_2024}, Nanonets-OCR2-3B~\cite{Nanonets2025OCR2}, GOT-OCR2~\cite{wei_got_2024}, PaddleOCR-VL~\cite{cui_paddleocr-vl_2025}, and DeepSeek-OCR~\cite{wei_deepseek-ocr_2025}.

\textbf{Task format.}
All models are evaluated in a zero-shot transcription setting. Each input consists of a single image and a task-neutral instruction equivalent to: ``convert the text and formulas in the image into Markdown, and do not include any other content.'' This prompt is intentionally short so that the evaluation measures visual transcription rather than task-specific reasoning or prompt engineering. No retrieval, task-specific demonstrations, or dataset-specific context is provided. Open-source models are served through OpenAI-compatible local inference services when possible, while closed-source models are evaluated through their official APIs. We use deterministic decoding settings when supported by the model interface.

\textbf{Normalization and tokenization.}
We apply the same post-processing to every model output, following the released evaluation script. Markdown code fences and explicit format wrappers such as \texttt{plaintext}, \texttt{markdown}, and \texttt{latex} are removed before scoring. To prevent runaway generations from dominating edit-distance metrics, single-line formula outputs are capped at 128 characters and other outputs are capped at 1024 characters before metric computation. For English text, line breaks are normalized to spaces and outputs are tokenized at the word level. For Chinese text, line breaks are removed and Jieba tokenization is used for token-level metrics. For public formula subsets, \LaTeX\ delimiters and non-semantic whitespace are normalized and tokens are extracted with a command-aware regular expression. For private mixed Chinese-formula subsets, we use a hybrid tokenizer that separates Chinese spans from \LaTeX{}-style commands, numbers, operators, and punctuation. No model-specific correction, manual cleanup, or outlier removal is applied during final scoring.

\textbf{Evaluation safeguards.}
The protocol is intentionally conservative. We use a single task-neutral instruction for all systems to avoid tuning prompts for particular model families or subsets. We also avoid semantic correction during scoring: an output that repairs a writer error, changes an incorrect formula into a correct one, or adds a missing reasoning step is penalized if the content is not visually supported by the image. The output-length caps do not change ordinary transcriptions, but they prevent pathological generations from overwhelming edit-distance metrics and making a small number of failures dominate aggregate scores. These safeguards make the evaluation closer to an offline benchmark of visual transcription than to a test of problem-solving or answer reconstruction.

\subsection{Metrics}
We employ five complementary metrics and report all values as percentages. BLEU-4~\cite{papineni_bleu_2001} measures 4-gram overlap between the model output and the ground truth. Token-level F1~\cite{f1_score} balances precision and recall, rewarding outputs that include the correct tokens without excessive insertion. CER and WER~\cite{cer_wer} measure character-level and token-level edit rates, respectively, so lower values indicate fewer transcription errors. Finally, 1-NED~\cite{ed} converts normalized Levenshtein distance into an accuracy-style score, with 1 indicating an exact match. We treat 1-NED as the primary metric because it is robust across text and formula settings while still penalizing substitutions, deletions, and hallucinated insertions. In tables with aggregate columns, \textit{Overall}~$\uparrow$ is the arithmetic mean of BLEU-4, F1, and 1-NED, while \textit{Overall}~$\downarrow$ is the arithmetic mean of CER and WER.

\begin{figure*}[tbp]
    \centering
    \includegraphics[width=0.48\linewidth]{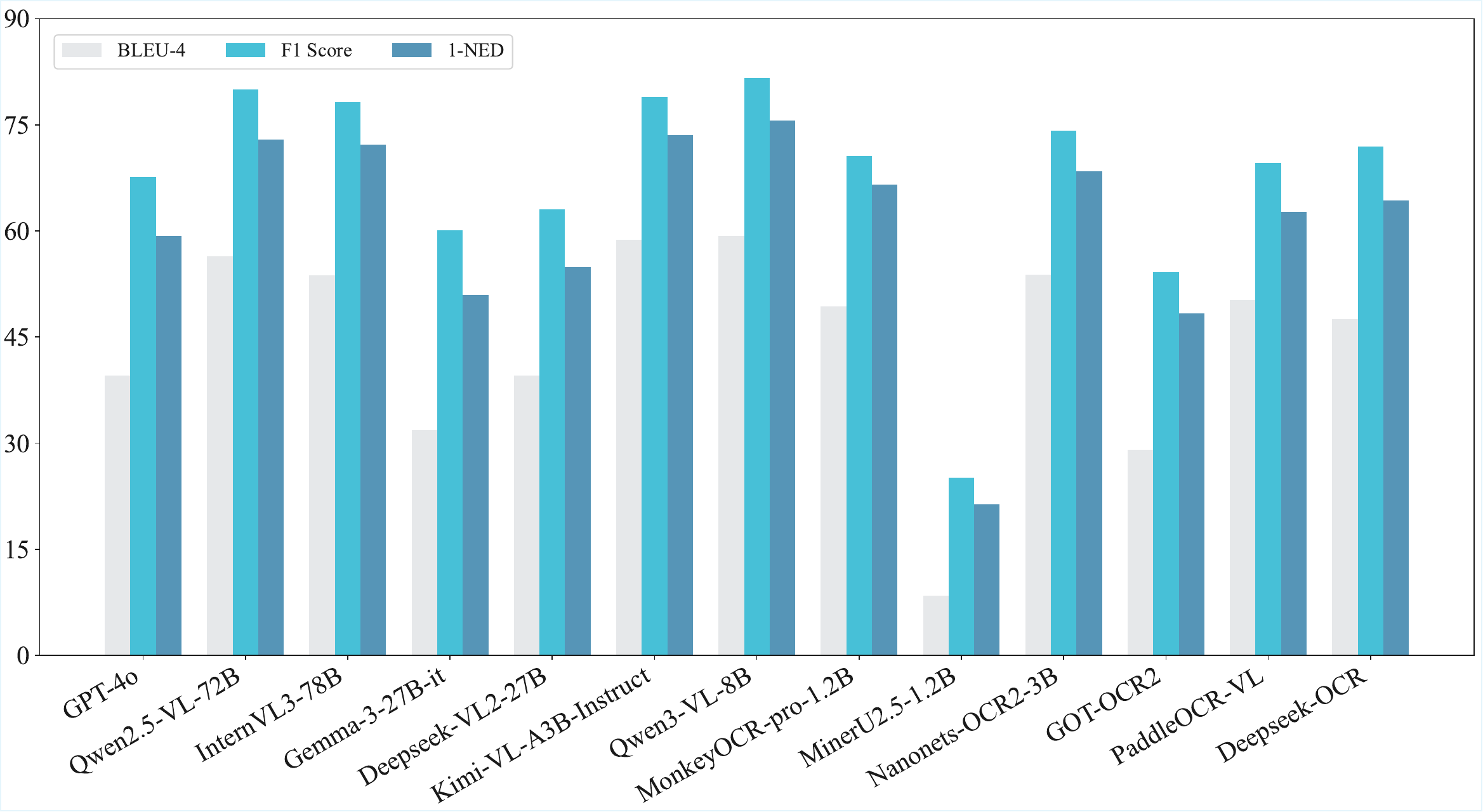}
    \includegraphics[width=0.48\linewidth]{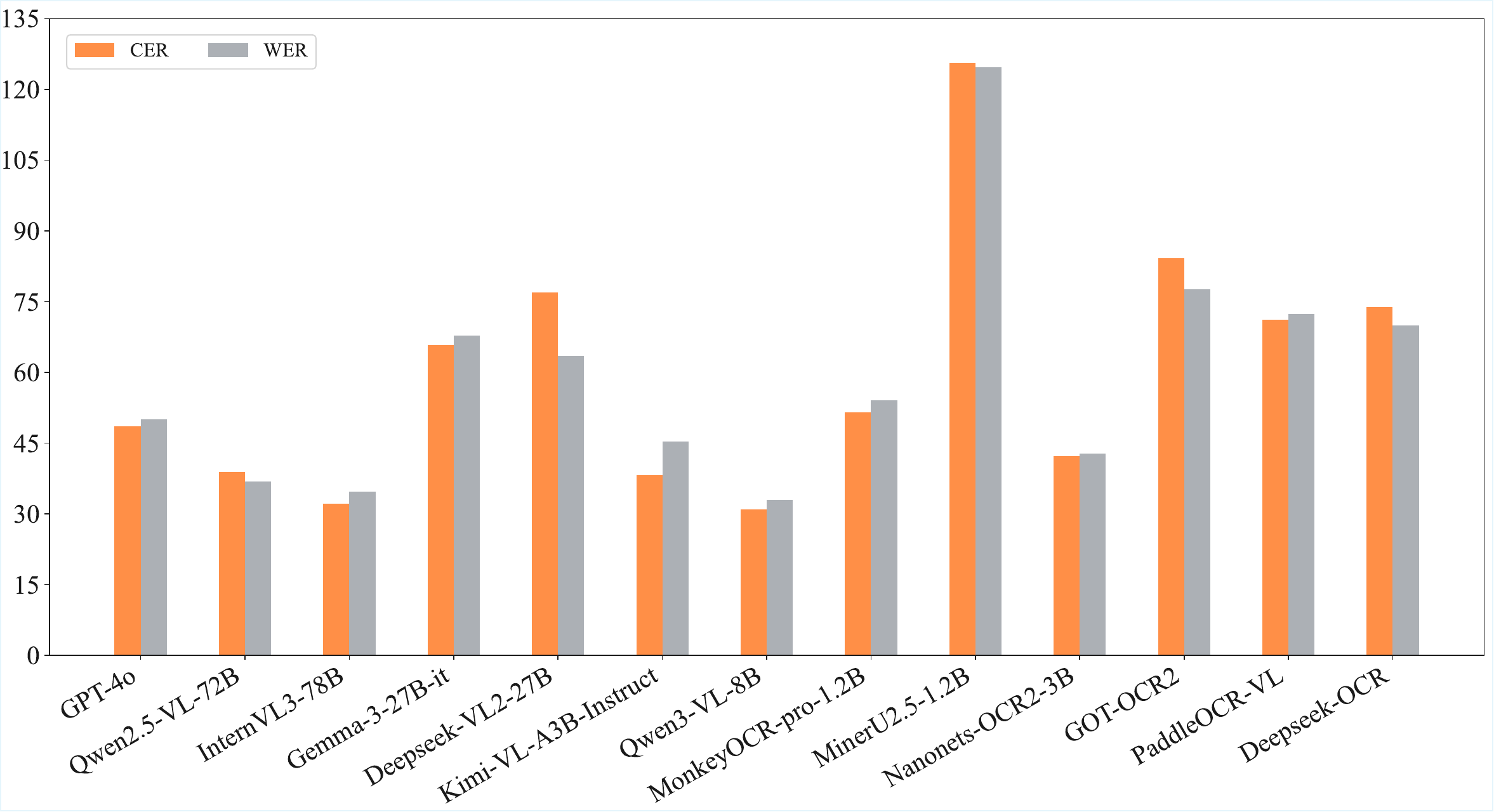}
    \vspace{-2mm}
    \caption{Overall benchmark performance (\%) under BLEU-4, F1, 1-NED, CER, and WER.}
    \label{fig:overall_1}
\vspace{-3mm}
\end{figure*}

\subsection{Main results}

Table~\ref{tab:performance_on_all} and Figure~\ref{fig:overall_1} summarize overall performance. Qwen3-VL-8B obtains the strongest aggregate score among all evaluated systems, with 59.24 BLEU-4, 81.60 F1, 75.63 1-NED, and the lowest average error rate among the MLLMs. However, the aggregate ranking also shows that handwritten OCR is far from solved: even the best model leaves a large gap to exact transcription, and several specialized OCR systems remain competitive on particular metrics. Among OCR models, Nanonets-OCR2-3B achieves the best overall OCR-model score and lower CER/WER than other specialized OCR baselines, indicating that domain-specific OCR training still provides advantages for precise transcription.

A closer comparison across model groups shows that overall handwritten OCR performance is not determined by model size alone. Qwen3-VL-8B outperforms larger MLLMs such as Qwen2.5-VL-72B and InternVL3-78B in the aggregate table, while Kimi-VL-A3B-Instruct later becomes strongest on several formula subsets. The OCR baselines also show substantial variation: Nanonets-OCR2-3B and MonkeyOCR-pro-1.2B remain competitive, whereas some document-oriented OCR models degrade sharply on handwritten formula content. These results suggest that training data, visual resolution handling, and formula-aware generation matter as much as general multimodal capacity for handwritten OCR.

The per-language results in Table~\ref{tab:performance_on_en_zh} show that model behavior is strongly language dependent. The Qwen series is consistently strong on Chinese handwriting, with Qwen2.5-VL-72B reaching 10.42 CER and 15.16 WER on zh-text. In contrast, Nanonets-OCR2-3B leads most English-text metrics among OCR models, while PaddleOCR-VL achieves the highest F1 and 1-NED on Chinese text within the OCR group. These results indicate that a single overall handwritten OCR score can obscure language-specific strengths and weaknesses.

Formula recognition further changes the ranking, as shown in Tables~\ref{tab:performance_on_sl_ml_easy} and~\ref{tab:performance_on_ml_medium_hard}. Kimi-VL-A3B-Instruct is the strongest model on single-line and easy multi-line formulas, reaching 93.33 F1 and 10.91 CER on single-line formulas. As the multi-line formula subset becomes harder, the leading model changes: Qwen3-VL-8B leads on ml-medium, while Qwen2.5-VL-72B is more robust on ml-hard. This rank instability confirms that OmniHandwritingOCR evaluates more than generic OCR ability; it exposes task-specific robustness under changing layout and structural complexity.

\subsection{Analysis and Findings}

\textbf{Structural complexity is the dominant stressor.}
All model families degrade as the benchmark moves from easy to hard multi-line formulas. For example, Qwen3-VL-8B drops from 86.92 F1 on ml-easy to 70.45 F1 on ml-hard, while its CER worsens from 24.95 to 57.08. This trend is consistent with the model-agnostic statistics in Table~\ref{tab:difficulty_validation}: average label length increases from 156.9 to 488.8 characters, average \LaTeX\ command count increases from 4.1 to 25.7, and average non-empty lines increase from 3.7 to 13.0. The degradation therefore cannot be explained only by harder handwriting styles. Hard samples require models to maintain long symbolic dependencies and preserve two-dimensional relationships such as fractions, superscripts, cases, and aligned derivations. The results suggest that current MLLM/OCR systems still struggle to convert complex handwritten spatial structures into faithful linear markup.

\textbf{Model rankings are task-conditioned.}
The strongest overall model is not uniformly strongest across all handwritten OCR scenarios. Qwen3-VL-8B leads the aggregate score, but Qwen2.5-VL-72B achieves the best Chinese-text CER/WER, Kimi-VL-A3B-Instruct dominates single-line and easy multi-line formulas, and Nanonets-OCR2-3B is highly competitive on English handwriting and OCR-model aggregates. This rank instability is important for evaluation: a single leaderboard score can hide whether a model is robust to language changes, formula structure, or long multi-line derivations. OmniHandwritingOCR is therefore designed to report both aggregate and task-conditioned performance.

\textbf{General-purpose and specialized models fail differently.}
General-purpose MLLMs tend to be stronger on broad cross-task aggregation, while specialized OCR models show sharper strengths on particular data regimes. Specialized OCR systems often produce more compact transcriptions, but may fail on mixed natural-language and formula content or on long reasoning-style derivations. Conversely, MLLMs can leverage language priors and broad visual understanding, but those same priors can introduce unsupported corrections when the image is ambiguous. This contrast suggests that handwritten OCR evaluation should measure not only recognition accuracy, but also whether the output remains visually grounded.

\textbf{Generative recognition introduces hallucinated correction.}
The fact-based labels allow us to observe a failure mode that is difficult to measure in ordinary OCR datasets: models may correct, complete, or fabricate content rather than transcribe what is visually present. We observe four recurring error patterns: correcting writer mistakes, inserting plausible intermediate steps, dropping visually present but hard-to-read symbols, and drifting into Markdown or explanatory formatting. This behavior is reflected quantitatively by error rates above 100 in some settings, such as DeepSeek-VL2-27B's 133.85 CER on Chinese handwriting and DeepSeek-OCR's 154.81 CER on single-line formulas. Such values mean that insertions dominate the edit distance, which is consistent with over-generation or hallucinated transcription. Figure~\ref{fig:label} further illustrates cases where models correct students' original mistakes, producing semantically plausible outputs that are wrong under a faithful-recognition criterion.

\textbf{Implications for knowledge processing.}
These findings are important beyond OCR accuracy itself. In retrieval, indexing, educational analytics, and document understanding systems, recognized text is often stored and reused as if it were a faithful representation of the source image. Deletions can remove searchable evidence, substitutions can corrupt formulas or named entities, and hallucinated corrections can introduce facts that were never written. The benchmark therefore highlights a practical requirement for handwritten OCR in knowledge pipelines: systems should preserve visual evidence even when the image contains mistakes or incomplete reasoning. This also explains why aggregate scores are insufficient. A model that performs well on clean text may still be unreliable for long mathematical derivations or mixed handwriting, where downstream systems need traceable transcription rather than fluent reconstruction.

\section{Data Release, Ethics, and Limitations}

\textbf{Data release and reproducibility.}
OmniHandwritingOCR is designed as an offline benchmark with deterministic scoring once model outputs are generated. We will publicly release the benchmark images, labels, and split manifests, with anonymized identifiers for newly collected educational data. We will also release the evaluation scripts that implement the normalization, tokenization, output-length caps, and metric computation described above. Public-source subsets follow their original dataset licenses, and all released private samples are anonymized to remove personally identifying information.

\textbf{Ethics.}
The private handwriting samples come from educational scenarios and are used only for recognition evaluation. Personally identifying metadata is removed, and the benchmark does not require demographic labels. Our fact-based annotation policy intentionally preserves visible writer errors; this design is important for evaluating faithful transcription, but it should not be interpreted as judging the writer's ability or correctness.

\textbf{Limitations.}
The benchmark focuses on offline image-based handwriting and does not include online pen trajectories. Its formula scoring is based on \LaTeX{}-style textual comparison, which can penalize semantically equivalent expressions written in different markup forms. The difficulty score includes a model-performance component, so it should be viewed as an evaluation stratification tool rather than an intrinsic human difficulty measure; the model-agnostic statistics in Table~\ref{tab:difficulty_validation} are included to show that the splits also reflect observable structural complexity. In addition, some public subsets may have appeared in the pretraining data of large multimodal models, which is a common concern for public benchmark evaluation. The newly collected private educational samples therefore provide a less exposed stress test for current systems. Future extensions can add human-only difficulty annotations and structure-aware formula equivalence metrics.

\section{Conclusion}

We introduced OmniHandwritingOCR, a diagnostic benchmark and unified evaluation protocol for handwritten OCR in MLLM/OCR systems. By integrating multilingual handwritten text with single-line and difficulty-stratified multi-line formulas, the benchmark tests capabilities that are underrepresented in existing OCR evaluations. Our evaluation of thirteen systems shows that aggregate performance remains limited, model rankings vary across language and formula settings, and structural complexity causes substantial degradation even for the strongest models. The fact-based annotation policy further exposes hallucinated correction, a failure mode that is especially important for generative OCR systems. These findings suggest that robust handwritten OCR for knowledge extraction requires evaluation protocols that are task-conditioned, structure-aware, and grounded in faithful transcription rather than plausible correction.

\section*{GenAI Usage Disclosure}
Generative AI tools were used in this research in two roles. First, vision-language models were used for pre-annotation as described in Section~\ref{sec:benchmark}. Second, GenAI-based coding assistants were used to assist with code development for conducting the experiments.

\clearpage
\bibliographystyle{ACM-Reference-Format}
\bibliography{main}

\end{document}